\documentclass[twocolumn]{article}

\usepackage[english]{babel}
\usepackage[pdftex]{graphicx}
\usepackage{color}
\usepackage{amsmath}
\usepackage{amssymb}
\usepackage{longtable}
\usepackage{url}
\usepackage{enumerate}
\usepackage{algorithm}
\usepackage{algorithmic}
\usepackage[pdftex,colorlinks=true,citecolor=black,
           pagecolor=black,linkcolor=black,menucolor=black,
           urlcolor=black]{hyperref}
\usepackage{afterpage}

\usepackage{natbib}
\usepackage{tikz,pgfplots}
\pgfplotsset{
compat=newest
}
\usepackage{comment}
\usepackage[caption=false]{subfig}

\title{Approximate Inference for Nonstationary Heteroscedastic \\ Gaussian process Regression}
\author{
    Ville Tolvanen\\
    Department of Biomedical Engineering and \\ Computational Science\\
    Aalto University\\
    ville.tolvanen@aalto.fi
  \and
    Pasi Jylänki\\
    Donders Institute for Brain, Cognition, and Behavior \\
    Radboud University Nijmegen \\
    pasi.jylanki@ru.nl
    \and
    Aki Vehtari\\
    Department of Biomedical Engineering and \\ Computational Science\\
    Aalto University\\
    aki.vehtari@aalto.fi
}
\date{\today}

\begin{document}

\maketitle

\begin{abstract}
   This paper presents a novel approach for approximate integration over the uncertainty of noise and signal variances in Gaussian process (GP) regression. Our efficient and straightforward approach can also be applied to integration over input dependent noise variance (heteroscedasticity) and input dependent signal variance (nonstationarity) by setting independent GP priors for the noise and signal variances. We use expectation propagation (EP) for inference and compare results to Markov chain Monte Carlo in two simulated data sets and three empirical examples. The results show that EP produces comparable results with less computational burden.   
\end{abstract}

\renewcommand{\v}[1]{{\bf #1}}
\newcommand{\vs}[1]{{\boldsymbol #1}}
\newcommand{\N}[0]{\text{N}}
\newcommand{\dd}{\,\mathrm{d}}

\section{Introduction}

Gaussian processes \citep[GP,][]{gpkirja} are commonly used as flexible non-parametric Bayesian
priors for functions. They provide an analytical framework that can be applied to various probabilistic learning tasks, for example, in geostatistics, gene expression time series \citep{hensman2013}, and density estimation \citep{riihimaki2014}.   
A typical assumption is that the parameters of the GP model stay constant
over the input space. However, this is not reasonable when it is
clear from the data that the phenomenon changes over the input space \citep[see, \emph{e.g.},][]{silverman1985}. 

As an improvement to these cases, \citet{goldberg1997} proposed heteroscedastic noise inference for Gaussian processes using a second GP to infer the log noise variance and doing the inference by Markov chain Monte Carlo (MCMC). More recent work on heteroscedastic noise models include solving the problem by transformation of the mean and variance parameters to natural parameters of Gaussian distribution \citep{le2005}, considering a two-component noise model \citep{naish2007}, and an expectation maximization like algorithm \citep{kersting2007}.   
\citet{adams2008} used expectation propagation \citep[EP,][]{minka2001a,minka2001b} to the model input-dependent signal variance (signal magnitude) in GPs by factoring the output signal to a product of a strictly positive modulating signal and a non-restricted signal, with independent GP priors for both of the signals. 

Non-stationarity can also be incorporated to the length-scales as proposed by \citet{gibbs1997} and further developed by \citet{paciorek2004}, where both used MCMC for the approximative inference. In general, the length-scale and the signal variance of a GP are underidentifiable and the proportion of them is more important
to the predictive performance \citep{Diggle+Tawn+Moyeed:1998,Zhang:2004,Diggle+Ribeiro:2007}. Therefore, we
assume that a GP with input-dependent signal variance and a GP with input-dependent length-scale would produce similar predictions. Thus, in this paper we concentrate on the input-dependent signal variance.

In this work, we present a straightforward and fast approach to integration over the uncertainty of the noise and signal variance in GP regression using EP. This approach can also be applied to input-dependent noise and signal variance by giving them independent GP priors. We extend the heteroscedastic noise model by \citet{goldberg1997} to EP inference, and extend the nonstationary model by \citet{adams2008} to analytical predictions. We consider the joint posterior of the modulating signal and the non-restricted signal and show that modeling the posterior correlations leads to significant improvements in the convergence of the EP algorithm compared to the factorized approximation. We also obtain stable analytical gradients of the log marginal likelihood.

We still need to infer other covariance function parameters such as the characteristic length-scale by  maximizing the marginal likelihood or posterior density, or using quadrature or MCMC integration. The performance of the EP implementation is compared to full MCMC \citep{neal1998} which produces the exact solution in the limit of an infinite sample size. We also compare the EP approximation of the latent posterior to an MCMC approximation, where we sample only the posterior of the latent values but use the EP-optimized hyperparameters. 

This paper is structured as follows. In Section~\ref{GPR} we briefly go through Gaussian process regression. Section~\ref{approximate_inference} is dedicated to the models and methods including the EP algorithm for posterior approximation, marginal likelihood evaluation and predictions. The experiments in Section~\ref{experiments} present the performance of our EP approach in two simulated data sets and three empirical problems. Finally the methods and results are discussed in Section~\ref{discussion}. 

\section{Gaussian Process Regression}
\label{GPR}
In standard GP regression the output $y$ is modeled as a function $f$ plus some additive noise $\epsilon$ such that $y(\v x) = f(\v x) + \epsilon$.
If $\epsilon \sim  \text{N}(0, \sigma^2)$, $y$ can be expressed as
\begin{equation}
y(\v x) \sim \N(f(\v x), \sigma^2).
\label{yn}
\end{equation}
The function $f$ is given a Gaussian process prior,
\begin{equation}
	f(\v x) \sim \mathcal{GP}(m(\v x), k(\v x, \v x')),
\label{GP}
\end{equation}
defined by its mean and covariance functions. In this work we use 
zero mean Gaussian processes for notational convenience. As for the covariance function, 
we use the common squared exponential (exponential quadratic):
\begin{equation}
	k(\v x, \v x') = \sigma^2_f \exp \left( -\sum_{i=1}^d\frac{(x_i - x_i')^2}{2\ell_i^2} \right),
	\label{sexp}
\end{equation}

where $\v x, \v x' \in \mathbb{R}^d$,  $\sigma^2_f$ is the magnitude or signal variance of the covariance function and $\ell_i$ is the characteristic length-scale corresponding to the $i$th input dimension. 

Given a data matrix $X = [\v x_1, \v x_2, \dots, \v x_n]$, we can write our GP prior for the latent function $f(\v x) = \v f$ as 
\begin{equation}
\v f \sim \text{N}(0, K(X,X)) = \text{N}(0, \v K_{\v f}),
\end{equation}
where the elements $[\v K_{\v f}]_{i,j} = k(\v x_i, \v x_j)$ are computed with \eqref{sexp}. 

In this work, we focus on models where either the noise variance in \eqref{yn}, or both the noise and signal variances in \eqref{sexp} depend on the input. These cases are handled analogously to \eqref{GP}, where the noise and signal variances are just some functions of the input, and the observation is combination of the three signals:
\begin{equation}
\begin{split}
  \log(\sigma^2 (\v x)) & \sim \mathcal{GP}(m(\v x), k_{\mathrm{n}}(\v x, \v x')), \\
  \log(\sigma^2_f(\v x)) & \sim \mathcal{GP}(m(\v x), k_{\mathrm{m}}(\v x, \v x')).
\end{split}
\end{equation}
From now on  $\vs \theta = \log (\sigma^2(\v x))$ and $\vs \phi = \log (\sigma^2_f(\v x))$. We set the GP prior for the logarithm of the variances to handle the positive restriction. 
We use the
squared exponential covariance function also for $k_\mathrm{n}(\v x, \v x')$ and
$k_\mathrm{m}(\v x, \v x')$, although other covariance functions could be used as well.

\section{Approximate Inference}
\label{approximate_inference}
In this section we go through the EP approximation, different models we use, and the algorithmic details.
\subsection{Expectation Propagation}
Expectation propagation is a general algorithm for forming an approximating distribution (from the exponential family) by matching the marginal moments of the approximating distribution to the marginal moments of the true distribution \citep{minka2001a, minka2001b}. The notation in this section follows mainly the notation of \citet{gpkirja}.

With Gaussian processes we wish to form the posterior distribution of the latent variables $\v f$ given the observations and inputs $p({\v f}\mid X,\v{y})$. However, the posterior distribution cannot be computed analytically in most cases, because the likelihood function and the prior distribution cannot be combined analytically. EP forms a Gaussian approximation to the posterior distribution by approximating the independent likelihood terms with Gaussian site approximations $\tilde{t}_i$. This enables the analytical computation of the posterior distribution because both the likelihood approximation and the prior are Gaussian:
\begin{equation}
p(y_i \mid f_i) \simeq \tilde{Z}_i \tilde{t}_i(f_i) = \tilde{Z}_i \text{N}(f_i \mid \tilde{\mu}_i, \tilde{\Sigma}_i),
\end{equation}
where $\tilde{Z}_i, \tilde{\mu}_i$ and $\tilde{\Sigma}_i$ are the parameters of the site approximations, or \emph{site parameters}. 
We use EP to approximate the posterior of $\v f$ such that
\begin{align}
& p({\v f} \mid X, {\v y}) = \frac{1}{Z}p({\v f} \mid X) \prod_i p(y_i \mid f_i) \nonumber \\ 
&\quad \approx \frac{1}{Z_{\mathrm{EP}}}p({\v f} \mid X) \prod_i \tilde{t}_i(f_i) = q({\v f} \mid X, {\v y}),
\end{align}
where $Z$ is the normalization constant or \emph{marginal likelihood},  $Z_{\mathrm{EP}}$ is the EP approximation to the marginal likelihood, $p({\v f} \mid X)$ is the prior of the latent variables $\v f$, and $q({\v f} \mid X, {\v y})$ is the Gaussian approximation to the exact posterior distribution $p({\v f} \mid X,{\v y})$.

\subsection{Noise Variance}
To integrate over the uncertainty of the noise variance in GP regression, we approximate the Gaussian likelihood as a product of two independent Gaussian site approximations $\tilde{t}_i$ for the mean $f_i$ and for the logarithm of the noise variance $\theta$:
\begin{align}
p(y_i \mid f_i, \sigma^2) &= \text{N}(y_i \mid f_i, \sigma^2) = \text{N}(y_i \mid f_i, e^{\theta}) \nonumber \\
&\approx \tilde{Z}_i \tilde{t}_i(f_i) \tilde{t}_i(\theta).
\end{align}
The posterior approximation of the latent variables $\v f$ and $\theta$ can now be written in a factorized form, if we set an independent prior distributions for $\v f$ and $\theta$
\begin{equation}
	p(\v f, \theta \mid X,\v y) \approx q(\v f \mid X,\v y) \, q(\theta \mid X,\v y).
\end{equation}
\subsection{Signal Variance}
If we wish to use the same approach as for noise variance to also integrate over the uncertainty of the signal variance, we need to move the signal variance from the GP prior to the likelihood function. Otherwise we would need to integrate over an $n$-by-$n$ matrix determinant, which is computationally expensive. To move the signal variance to the likelihood function, we reparameterize $\v f$ as $\v f = \sigma_f \tilde{\v f}$, where $\sigma_f$ is the square root of the signal variance. Now, if $\mathrm{Cov}[\v f] = \sigma_f^2 K$, then $\mathrm{Cov}[\tilde{\v f}] = K$, where $K$ is covariance matrix computed with identity signal variance in \eqref{sexp}. As noted in Section~\ref{GPR}, we model the logarithm of the signal and noise variances to take into account the restriction for them to be positive.  
Because both $\tilde{\v f}$ and $\phi$ model the mean of the distribution, we expect them to have strong correlation. Thus, instead of doing a factorized approximation as for the noise variance, we approximate the likelihood with two site approximations: one for the noise variance and a joint two-dimensional Gaussian for $\vs v_i = (\tilde{f}_i, \phi)$:
\begin{align}
p(y_i \mid \tilde{f}_i, \theta, \phi) &= \text{N}(y_i \mid e^{\phi/2}\tilde{f}_i, e^{\theta})  \nonumber \\
  &\approx \tilde{Z}_i \tilde{t}_i(\tilde{f}_i, \phi)\, \tilde{t}_i(\theta).
\end{align}
Assuming independent priors for the latent variables $\tilde{\v f}$, $\phi$ and $\theta$, the posterior approximation is also analogous to the noise variance case, such that
\begin{equation}
p(\tilde{\v f}, \theta, \phi \mid X,\v y) \approx q(\tilde{\v f}, \phi \mid X,\v y)\, q(\theta \mid X,\v y).
\end{equation}
It should be noted that we also tested the fully factorized approximation $\tilde{t}_i(\tilde{f}_i, \phi) = \tilde{t}_i(f_i)\tilde{t}_i(\phi)$, but it gave worse predictions, and the EP algorithm needed clearly more iterations to converge.
\subsection{Input-Dependent Noise and Signal Variance}

We can easily extend the presented likelihood approximations to also include input-dependency on signal and noise variances (or either one), by setting independent GP priors for both the logarithm of the noise variance and logarithm of the signal variance:
\begin{equation}
\begin{split}
p(\vs \theta \mid X) &= \text{N}(0, \v K_{\vs \theta}), \\
p(\vs v \mid X) &= \text{N}(0, \v K_{\vs v}).
\end{split}
\end{equation}
If we integrate over the input-dependent signal variance, we have
\begin{equation}
\v K_{\vs v} =
\begin{bmatrix}
\v K_{\tilde{\v f}} & 0 \\
0 & \v K_{\vs \phi}
\end{bmatrix},
\end{equation}
otherwise we have $\v K_{\vs v} = \v K_{\v f}$. The covariance matrices are computed from the squared exponential covariance function \eqref{sexp}. 
By setting the GP priors, we assume that the signal and noise variances are also some unknown functions that depend on the input $\v x$. 

The site approximations are of the same form independent of the input-dependency of the parameters
\begin{equation}
\begin{split}
  \tilde{t}_i(\theta_i) &= \text{N}(\tilde{\mu}_{\theta,i}, \tilde{\Sigma}_{\theta,i}), \\
  \tilde{t}_i(\vs v_i) &= \text{N}(\tilde{\vs \mu}_{v,i}, \tilde{\vs \Sigma}_{v,i}).
\end{split}
\end{equation}
If we integrate over the (input-dependent) signal variance, we have
\begin{equation}
  \tilde{\vs \mu}_{v,i} =
  \begin{bmatrix}
  \tilde{\mu}_{\tilde{f},i} \\
  \tilde{\mu}_{\phi,i}
  \end{bmatrix}
  \quad \text{and} \quad
  \tilde{\vs \Sigma}_{v,i} = 
  \begin{bmatrix}
  \tilde{\Sigma}_{\tilde{f},i} & \tilde{\Sigma}_{\tilde{f}\phi,i} \\
  \tilde{\Sigma}_{\tilde{f}\phi,i} & \tilde{\Sigma}_{\phi,i}
  \end{bmatrix},
  \label{sitev}
\end{equation}
otherwise $\tilde{\vs \mu}_{v,i} = \tilde{\mu}_{f,i}$ and $\tilde{\vs \Sigma}_{v,i} = \tilde{\Sigma}_{f,i}$. Here we have used $\Sigma$ for both the scalar variance of the univariate Gaussian and the covariance matrix of the bivariate Gaussian, but it should be clear from the context which one it represents.

The posterior distributions can be computed with
\begin{align}
q(\vs v \mid X, \v y) &= \text{N}(\vs \mu_{\vs v}, \vs \Sigma_{\vs v}) \propto p(\vs v \mid X) \prod_i \tilde{t}_i(\vs v_i) \nonumber \\ 
&= \text{N}(\vs v \mid 0, \v K_{\vs v}) \text{N}(\vs v \mid \tilde{\vs \mu}_{\vs v}, \tilde{\vs \Sigma}_{\vs v}), \label{vposterior} \\
q(\vs \theta \mid X, \v y) &= \text{N}(\vs \mu_{\vs \theta}, \vs \Sigma_{\vs \theta}) \propto p(\vs \theta \mid X) \prod_i \tilde{t}_i(\theta_i) \nonumber \\
&= \text{N}(\vs v \mid 0, \v K_{\vs \theta}) \text{N}(\vs \theta \mid \tilde{\vs \mu}_{\vs \theta}, \tilde{\vs \Sigma}_{\vs \theta}) \label{thetaposterior},
\end{align}
where $\vs \mu_{\theta} = \vs \Sigma_{\theta} \tilde{\vs \Sigma}^{-1}_{\theta} \tilde{\vs \mu}_{\theta}$, $\vs \Sigma_\theta = (\v K_{\theta}^{-1} + \tilde{\vs \Sigma}_{\theta}^{-1})^{-1}$, 
$\vs \mu_{\vs v} = \vs \Sigma_{\vs v } \tilde{\vs \Sigma}^{-1}_{\vs v} \tilde{\vs \mu}_{\vs v}$, and 
$\vs \Sigma_{\vs v} = (\v K_{\vs v}^{-1} + \tilde{\vs \Sigma}_{\vs v}^{-1})^{-1}$. The joint site covariance $\tilde{\vs \Sigma}_\theta$ is diagonal while $\tilde{\vs \Sigma}_{\vs v}$ has a block form if we integrate over the input-depenedent signal variance
\begin{equation}
\tilde{\vs \Sigma}_{\vs v} = 
\begin{bmatrix}
\tilde{\vs \Sigma}_{\tilde{f}} & \tilde{\vs \Sigma}_{\tilde{f}\phi} \\
\tilde{\vs \Sigma}_{\phi\tilde{f}} & \tilde{\vs \Sigma}_{\phi} 
\end{bmatrix},
\end{equation}
where each block is diagonal. Cross-diagonal terms, $\tilde{\vs \Sigma}_{\tilde{f}\phi} = \tilde{\vs \Sigma}_{\phi\tilde{f}}$, collect the marginal covariances $\tilde{\Sigma}_{\tilde{f}\phi,i}$ and the main-diagonal terms, $\tilde{\vs \Sigma}_{\tilde{\v f}}$ and $\tilde{\vs \Sigma}_{\vs \phi}$, collect the marginal variances $\tilde{\Sigma}_{\tilde{f},i}$ and $\tilde{\Sigma}_{\phi,i}$. If we do not integrate over the signal variance, we have $\tilde{\vs \Sigma}_{\vs v} = \tilde{\vs \Sigma}_{f}$. 

\subsection{EP Algorithm}

The full EP algorithm is presented in Algorithm~\ref{epalgo}. The main points in the algorithm are the same as in the standard EP approach for Gaussian processes \citep[][pp. 52--60]{gpkirja}. However, there are some implementation details that should be noted:

1. The overall stability of the EP updates can be improved  by working in the natural parameter space of the site approximations. This means that we use the natural parameterization, $\tilde{\nu} = \tilde{\Sigma}^{-1}\tilde{\mu}$ and $\tilde{\tau} = \tilde{\Sigma}^{-1}$, for the site approximations. This way we can avoid inverting the site covariance matrices at every iteration. 

2. Even though the algorithm should be stable and robust, there are some cases where the site updates exhibit oscillations, for example, due to weird hyperparameter values in the covariance functions. Thus, the updates should be damped after computing the new site approximations in step~4, 
\begin{equation}
\begin{split}
& \quad \Delta \tau_i = \delta (\tau_i^{\text{new}} - \tau_i^{\text{old}}), \quad \Delta \nu_i = \delta (\nu_i^{\text{new}} - \nu_i^{\text{old}}) \quad  \\
& \quad \tau_i^{\text{new}} = \tau_i^{\text{old}} + \Delta \tau_i, \quad \nu_i^{\text{new}} = \nu_i^{\text{old}} + \Delta \nu_i,
\nonumber
\end{split}
\end{equation}
with some suitable damping factor $\delta$, for example $\delta = 0.8$. 

3. In step 3 of the algorithm we minimize KL divergence with respect to Gaussian distributions. This means that we match the first and second moments of the one-dimensional distributions and in addition to these the 
cross-moment if we have a bivariate Gaussian $\tilde{t}_i(\vs v_i)$. The integrals over $f_i$ or $\tilde{f}_i$ can be computed analytically in every case in steps 2 and 3. If we don't integrate over signal variance, this can be done trivially as both the cavity and likelihood are Gaussian with respect to $f_i$. If we integrate over signal variance, we can utilize the standard factorization of the multivariate Gaussian $q_{-i}(\tilde{f}_i, \phi_i) = q_{-i}(\tilde{f}_i\mid\phi_i)q_{-i}(\phi_i)$. The integrals over $\theta$ and $\phi$ must be computed numerically, but this can be done effectively, for example, with Simpson's method.

4. We use parallel EP updates for the site parameters. This means that we compute the site updates for every site approximations before we update the posterior distribution and compute the marginal likelihood. This usually results in a few more EP iterations than sequential EP, but the overall speed of the algorithm is faster.

\begin{algorithm}[h!]
\caption{Parallel EP algorithm}
\begin{algorithmic}\sffamily\footnotesize
\STATE
Initialize $\tilde{\mu}_{i,\theta}=\tilde{\mu}_{i,v}=\tilde{\Sigma}_{i,\theta}^{-1}=\tilde{\Sigma}_{i,v}^{-1}=0$ for $i=1,2,\dots, n$.

Set $q(\vs \theta \mid X,\v y) = p(\vs \theta \mid X)$ and $q(\vs v \mid X,\v y) = p(\vs v \mid X)$.
\REPEAT
\FOR{$i=1$ {\bfseries to} $n$}
\IF{input-dependent signal variance}
        \STATE $\vs v_i = (\tilde{f}_i, \phi_i)$
    \ELSE
        \STATE $\vs v_i = f_i$
    \ENDIF
\STATE
1. Compute the cavity distributions:
\begin{align}
\quad & q_{-i}(\vs v_i) \propto q_i(\vs v_i)/\tilde{t}_i(\vs v_i) \nonumber \\
\quad & q_{-i}(\theta) \propto q_i(\theta)/\tilde{t}_i(\theta) \nonumber
\end{align}
with
\begin{align}
\Sigma_{-i,\cdot}^{-1} &= \Sigma_{i,\cdot}^{-1}-\tilde{\Sigma}_{i,\cdot}^{-1}  \nonumber \\ 
\mu_{-i,\cdot} &= \Sigma_{-i,\cdot}(\Sigma_{i,\cdot}^{-1}\mu_{i,\cdot} - \tilde{\Sigma}_{i,\cdot}^{-1}\tilde{\mu}_{i,\cdot}), \nonumber
\end{align}
when
\begin{align}
\quad & q_i(\cdot) \sim \text{N}(\mu_{i,\cdot}, \Sigma_{i,\cdot}) \nonumber \\
\quad & \tilde{t}_{i}(\cdot) \sim \text{N}(\tilde{\mu}_{i,\cdot}, \tilde{\Sigma}_{i,\cdot}) \nonumber
\end{align}

2. Compute the normalization $\hat{Z}_i$:
\begin{equation}
\hat{Z}_i = \iint p(y_i \mid \vs v_i, \theta)q_{-i}(\vs v_i)q_{-i}(\theta) \dd \vs v_i \dd \theta \nonumber
\end{equation}
3. Find the best marginal posterior approximation for $q_i(\vs v_i)$ and $q_i(\theta_i)$ by 
\begin{align}
\quad &\min_{q_i(\vs v_i)} \textrm{KL}(\hat{Z}_i^{-1}p(y_i \mid \vs v_i, \theta)q_{-i}(\vs v_i)q_{-i}(\theta) \| q_i(\vs v_i)) \nonumber \\
\quad &\min_{q_i(\theta)} \textrm{KL}(\hat{Z}_i^{-1}p(y_i \mid \vs v_i, \theta)q_{-i}(\vs v_i)q_{-i}(\theta) \| q_i(\theta)). \nonumber
\end{align}
4. Update the site approximations $\tilde{t}_i$ by
\begin{align}
\tilde{t}_i(\vs v_i) &\propto q_i(\vs v_i)/q_{-i}(\vs v_i) \nonumber\\
\tilde{t}_i(\theta) &\propto q_i(\theta)/q_{-i}(\theta) \nonumber
\end{align}
analogously to step 1. 
\ENDFOR
\STATE
5. Update the posterior distributions with \eqref{vposterior}--\eqref{thetaposterior}.

6. compute the marginal likelihood with \eqref{Zep}.
\UNTIL{$Convergence$}
\end{algorithmic}
\label{epalgo}
\end{algorithm}

\subsubsection{Marginal Likelihood}
Marginal likelihood can be used for model selection under GP framework as it has good calibration and the maximum of the marginal likelihood usually corresponds to good predictions \citep{gpkirja, nickisch2008, riihimaki2013}. Marginal likelihood in Gaussian processes is defined as 
\begin{align}
Z &= p({\v y} \mid X) = \int p({\v f} \mid X) p({\v y} \mid {\v f})  \dd {\v f}.
\end{align}
For our noise and signal variance GPs, an EP approximation to the marginal likelihood is 
\begin{align}
& Z \approx Z_{\mathrm{EP}} = q({\v y} \mid X)  \nonumber \\
 &= \int p({\vs v} \mid X) p(\vs \theta \mid X) \prod_i \tilde{Z}_i \tilde{t}_i(\vs v_i)\tilde{t}_i(\theta_i)  \dd {\vs v} \dd \vs \theta,
\end{align}
where $\vs v = (\tilde{\v f}, \vs \phi)$ or $\vs v = \v f$.
Following \citet{cseke2011}, we define the term
\begin{align}
& \log Z({\v m}, {\v V}) = \frac{1}{2}{\v m}^\mathsf{T} {\v V}^{-1} {\v m} + \frac{1}{2}\log |\v V| + \frac{n}{2}\log (2\pi).
\label{Z}
\end{align}
Now the EP approximation for marginal likelihood can be computed with
\begin{align}
& \log Z_{\mathrm{EP}} = \log Z({\vs \mu}_\theta, {\vs \Sigma}_\theta) + \log Z({\vs \mu}_v, {\vs \Sigma}_v) \label{Zep} \nonumber \\
& \quad + \sum_i \left(\log Z(\mu_{-i,\theta}, \Sigma_{-i,\theta}) - \log Z(\mu_{i,\theta}, \Sigma_{i,\theta}) \right)  \nonumber \\
& \quad + \sum_{i} \left( \log Z(\mu_{-i,v}, \Sigma_{-i,v}) - \log Z(\mu_{i,v}, \Sigma_{i,v}) \right) \nonumber \\
& \quad - \log Z(0,K_{\vs v}) - \log Z(0,K_{\vs \theta}) + \sum_i \log \hat{Z}_i ,
\end{align}
where ${\vs \mu}$ and ${\vs \Sigma}$ are the parameters of the posterior distribution approximation $q(\cdot | X,{\v y})$, $\mu_i$ and $\Sigma_i$ are the $i$th marginal terms of ${\vs \mu}$ and ${\vs \Sigma}$, $\mu_{-i}$ and $\Sigma_{-i}$ are the $i$th marginal mean and variance parameters of the cavity distributions $q_{-i}(\cdot)$, and $K_j$ are the prior covariances from the GP.

Note that for $\theta$ the marginal parameters are one-dimensional, but for $\vs v$ they are two-dimensional if 
we integrate over the signal variance like for the site approximations in \eqref{sitev}.

\subsubsection{Predictions}

For predicting a future observation $y^*$ for input $\v x^*$, we need to compute the predictive distribution
\begin{align}
& p(y^* | \v x^*,X, \v y) = \iint p(y^*, \vs v^*, \theta^* | \v x^*, X, \v y) \dd \vs v^* \dd \theta^*  \nonumber \\
& =  \iint p(y^* | \vs v^*,\theta^*)q(\vs v^* | \v x^*,X,\v y)q(\theta^* | \v x^*,X,\v y) \dd \vs v^* \dd \theta^*, 
\end{align}
where 
\begin{equation}
	q(\vs v^* | \v x^*, X, \v y) = \int p(\vs v^* | \vs v)q(\vs v | X,y) \dd \v f
	\label{ppf}
\end{equation}
can be easily computed using properties of Gaussian processes. 
Note that if we assume stationary signal or noise variance, the respective posterior distributions reduce to one-dimensional Gaussian distributions. This means that $q(\vs v|X,\v y)$ becomes $n+1$ dimensional, and the posterior predictive distribution equals the posterior distribution. 
Because we approximate the posterior predictive distribution of the latent variables and the predictive distribution of $y^*$ by a Gaussian distribution, we can always compute the predictions analytically, regardless whether we have input-dependent signal or noise variance.
For a GP with EP marginalized noise variance we get the following predictive distributions
\begin{align}
& \mathbb{E}[y^* \mid \v x^*, X,{\v y}] = \mathbb{E}[f^* \mid \v x^*, X,{\v y}] \\ 
& \mathbb{V}[y^* \mid \v x^*, X,{\v y}] = \mathbb{V}[f^* \mid \v x^*, X,{\v y}] \nonumber \\
& + \exp \big(\mathbb{E}[\theta^* \mid \v x^*,X,{\v y}] + \frac{1}{2} \mathbb{V}[\theta^* \mid \v x^*, X, {\v y}] \big).
\end{align}
 For a GP with EP marginalized noise and signal variance the results are quite lengthy and are omitted here to save space (see supplementary material).

\begin{figure}[h!]
  \centering
  \pgfplotsset{
  	every axis y label/.style={at={(0,0.5)},xshift=-2em,rotate=90,anchor=center}
  }
  
  \subfloat[EP posterior approximations (contours) and the MCMC samples from the latent posterior.]{
    \newlength\figheight 
    \newlength\figwidth 
    \setlength\figheight{.45\columnwidth} 
    \setlength\figwidth{.45\columnwidth} 
    \footnotesize 
    \pgfplotsset{
    } 
    \hspace*{-0.2cm}%
    \input{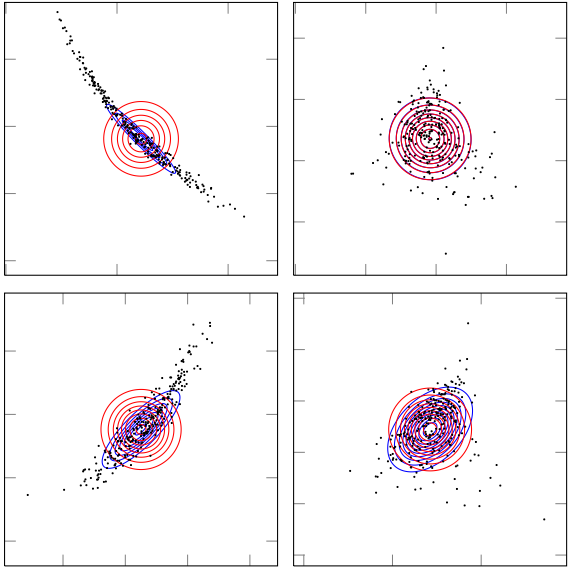}
  }
  \\
  \subfloat[Covergence of EP with factorized (red) and joint (blue) approximations.]{
    \setlength\figheight{.40\columnwidth} 
    \setlength\figwidth{.93\columnwidth} 
    \footnotesize \sffamily
    \pgfplotsset{
  	  every axis y label/.style={at={(0,0.5)},xshift=-2em,rotate=90,anchor=center},
  	  y tick label style={rotate=90}
    }
    \hspace*{-0.45cm}%
    \input{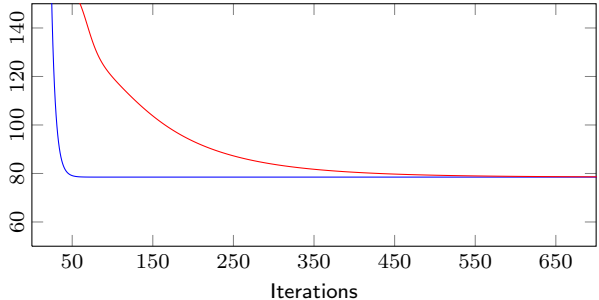}  
  }
  \caption{Example comparisons of EP posterior approximations with MCMC samples from the latent posterior and the covergence of the EP algorithm. Red contours correspond to the factorized approximation $q(\tilde{\v f} \mid X, \v y)q(\vs \theta \mid X, \v y)$ and the blue contours correspond to the full joint approximation $q(\tilde{\v f}, \vs \phi \mid X, \v y)$.}
  \label{factorized_comparison}
\end{figure}

\subsubsection{Factorized Approximation and Converge}

In this section we discuss certain key properties of the posterior
approximations introduced in Sections 3.2-3.4. More precisely, we illustrate
the importance of the utilized factorization assumptions in terms of both
accuracy and convergence of the resulting EP algorithm.

Panel (a) of Figure \ref{factorized_comparison} visualizes the marginal posterior
distributions of the latent values related to both the unscaled function
values $\tilde{f}_i$ (x-axis) and the magnitude process $\phi_i$ (y-axis).
Each of the four subplot shows the latent values associated with four
different observations (likelihood terms) resulting from a non-trivial
simulated data set (see Section~\ref{experiments}). MCMC samples from
the true posterior distribution are plotted with black dots together with two
different EP approximations: the partially coupled approximation $q(\tilde{\v{f}},
\vs \phi) q(\vs \theta)$ introduced in Section 3.4 (blue contours) and a fully
factorized approximation of the form $q(\tilde{\v f}) q(\vs \phi ) q(\vs \theta)$
(red contours).
Subplots on the left show strong posterior dependencies between the latent
values resulting from the combined effect of the within-observation couplings
$f_i = \tilde{f}_i \exp(\phi_i/2)$ and the between-observation correlations
controlled by the GP priors. On the other hand, subplots on the right show
much weaker couplings indicating that the the within-observation coupling
does not necessarily introduce strong posterior dependencies.
Comparison of the joint posterior approximations of $\theta_i$ with either
$\phi_i$, $\tilde{f}_i$, or $f_i = \tilde{f}_i \exp(\phi_i/2)$ did not show
strong dependencies, which is why we used a factorized approximation for
$\theta$ to facilitate computations.

According to our experiments, neglecting the posterior couplings does not
significantly affect the predictive performance compared to the
fully-factorized approximation. However, representing these couplings has a
significant effect on the convergence properties of the EP algorithm.
Subfigure (b) of Figure~\ref{factorized_comparison} shows the EP marginal 
likelihood approximations as a function
of EP iteration in both settings. The fully-coupled approximation (red line)
converges very slowly compared to the partially coupled approximation (blue
line); the former requires often hundreds of iterations whereas the
partially-coupled approach converges usually in less than 50 iterations. In
our experiments the convergence properties of the full-coupled algorithm
could not be improved by adjusting damping.

This behavior can be explained by slow propagation of information between the
latent values from different likelihood terms with the fully-coupled
approximation. Because each likelihood term is updated separately from the
others, information on the posterior dependencies in other site terms is not
available during the update.
These findings are fully congruent with the convergence differences in
multi-class GP classification when between-class dependencies are omitted
\citep{riihimaki2013}. 

\section{Experiments}
\label{experiments}

\begin{figure*}
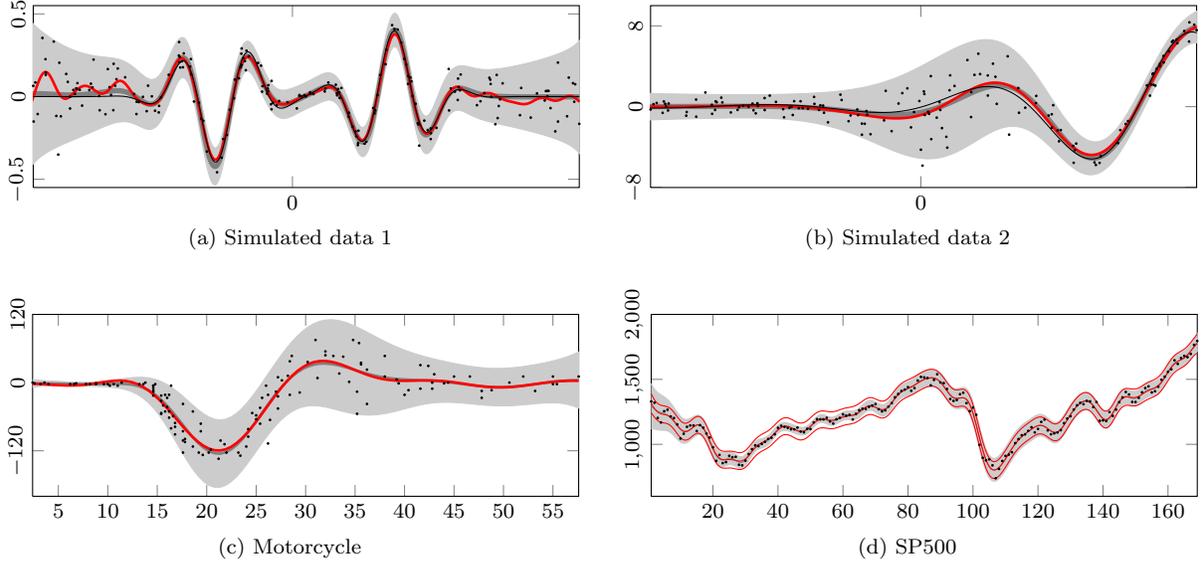

  \centering
  \pgfplotsset{
  	  every axis y label/.style={at={(0,0.5)},xshift=-2em,rotate=90,anchor=center},
  	  y tick label style={rotate=90}
  }  
  \setlength\figheight{.30\columnwidth} 
  \setlength\figwidth{.90\columnwidth} 
  \hspace*{\fill}%
  \subfloat[Simulated data 1]{
    \footnotesize 
    \input{toy1_fig2.tikz}
    \label{toy1_fig2}
  } \hspace*{\fill}
  \subfloat[Simulated data 2]{
    \footnotesize 
    \input{toy2_fig2.tikz}
    \label{toy2_fig2}
  }%
  \hspace*{\fill}
  \\
  \hspace*{\fill}%
  \subfloat[Motorcycle]{
    \footnotesize 
    \input{motorcycle_fig1.tikz}
    \label{motorcycle_fig1}
  }  \hspace*{\fill}
  \subfloat[SP500]{
    \footnotesize 
    \input{sp500_fig1.tikz}
    \label{sp500_fig1}
  }%
  \hspace*{\fill}%
  \caption{One-dimensional data sets and the EP predictions with uncertainty intervals. Thin black lines correspond to the true signal in the simulated data sets, and the thick gray lines are the GP predictions with EP. The grey area is the 95\% credible interval of the prediction. Red lines correspond to the standard GP prediction with MAP values for the signal and noise variance (credible intervals only shown for SP500).}
  \label{toy_data}
\end{figure*}

In this section we go through the different data sets we use for experiments, different methods and the assessment criteria for the results.

{\bf Simulated data 1}. The first simulated data was generated by the following setup:
\begin{equation}
\begin{split}
	\tilde{f}(x) &= \sin(x), \\
	\sigma_f(x) &= \textrm{N}(x \mid -2.5, 1) + \textrm{N}(x \mid 2.5, 1), \\ 
	\sigma(x) &= 0.08+\textrm{N}(x \mid -8,3)+\textrm{N}(x \mid 8,3),  \\ 
		y(x) &= \sigma_f(x) \tilde{f}(x) + \epsilon,
\end{split}
\end{equation}
where $\epsilon \sim \textrm{N}(0, \sigma(x))$. The training data was generated by first drawing 200 random $x$ values from $\mathrm{U}(-8,8)$. After this we computed the mean signal by combining the modulating signal $\sigma_f(x)$ and $\tilde{f}(x)$. Then some random noise with standard deviation $\sigma(x)$ was added. For the test set we used uniform grid of 1000 points in the interval $(-8,8)$ and computed the function values analogously to training set, without adding noise. The experiment was repeated 100 times for different realizations of the training data set to assess the variation in the final predictions of the test set.

\noindent {\bf Simulated data 2}. The second simulated dataset was generated with  
\begin{equation}
\begin{split}
	\tilde{f}(x) &= \sin(x), \\
	\sigma_f(x) &= \exp(2\sin(0.2x)), \\ 
	\sigma(x) &= \exp(0.75\sin(0.5x+1))+0.1, \\
		y(x) &= \sigma_f(x) \tilde{f}(x) + \epsilon.
\end{split}
\end{equation}
The training and test data were generated analogously to the first experiment. We used 150 training points and the different generating signals for the observations. The second experiment was also repeated 100 times as in the first experiment.  

\noindent{\bf Motorcycle}. The motorcycle data  \citep{silverman1985} consists of 133 accelerometer readings in a simulated motorcycle crash.

\noindent {\bf Concrete}. The second empirical experiment uses concrete quality data \citep{vehtari2002, jylanki2011}, where the output is volume percentage of air in concrete, air-\%, with 27 different input variables.
The input variables depend on the properties of the stone materials, additives and the amount of cement and water.

\noindent {\bf SP500}. The last empirical experiment is concerned with predicting the SP500 index. The data set consists of monthly averages of the index between years 2001--2014, with a total of 169 observations. We demonstrate on this data how a GP with input-dependent noise variance works as a stochastic volatility model.

\begin{table*}
\begin{center}
\caption{The table shows MLPD values for different methods, where higher values correspond to better predictions. For the  concrete data \emph{ISO} means that we have an isotropic covariance functions for all the latent variables, and \emph{ARD} denotes automatic relevance determination for $\v f$ and $\tilde{\v f}$, and an isotropic covariance function for the rest of the latent variables. }
\vspace{0.5cm}
\begin{tabular}{l|*{6}{c}}
Method              & Simulated 1 & Simulated 2 & Motorcycle & Concrete (ISO) & Concrete (ARD) & SP500\\
\hline
 GP					& 0.95 $\pm$ 0.026 & $-1.70 \pm 0.034$ & $-0.71$ & 0.06 & 0.11& 0.27\\
 EP (n)	    	    & 1.22 $\pm$ 0.025 & $-1.49 \pm 0.032$ & $-0.41$ & 0.13 & 0.21& 0.42\\
 EP (m+n)   		& 1.23 $\pm$ 0.028 & $-1.47 \pm 0.029$ & $-0.42$ & 0.22 & 0.26& 0.41\\
 EP-MC (n)    		& 1.22 $\pm$ 0.025 & $-1.49 \pm 0.032$ & $-0.40$ & 0.11 & 0.23& 0.43 \\
 EP-MC (m+n)    	& 1.24 $\pm$ 0.023 & $-1.47 \pm 0.029$ & $-0.41$ & 0.21 & 0.28& 0.42 \\
 MCMC    			& 0.95 $\pm$ 0.020 & $-1.70 \pm 0.025$ & $-0.71$ & 0.07 & 0.13& 0.28 \\
 MCMC (n)   		& 1.22 $\pm$ 0.021 & $-1.55 \pm 0.150$ & $-0.39$ & 0.10 & 0.22& 0.19 \\
 MCMC (m+n)   		& 1.24 $\pm$ 0.019 & $-1.49 \pm 0.030$ & $-0.40$ & 0.20 & 0.19& 0.26 \\ \hline
\end{tabular}
\label{results}
\end{center}
\end{table*}

We compare 8 different methods: {\bf GP} (Standard GP regression), {\bf EP(n)} and {\bf MCMC(n)} (integration over input-dependent noise variance with EP and MCMC), {\bf EP(n+m)} and {\bf MCMC(m+n)} (integration over input-dependent signal and noise variance with EP and MCMC), {\bf EP-MC(n)} and {\bf EP-MC(m+n)} (EP optimized hyperparameters for covariance functions and sampling of the posterior of the latent variables). 

Figure~\ref{toy_data} presents the behaviour of the EP (m+n) for the one-dimensional experiments. 

In standard GP regression we use \emph{maximum a posteriori} (MAP) values for all the model parameters (signal variance, noise variance, length-scales). In the EP methods, when integrating over input-dependent noise variance, we use MAP values for signal variance and length-scales, and when integrating over input-dependent signal and noise variance, we use MAP values for the length-scales. Latent MCMC means that we use EP optimized MAP values for the covariance function parameters and sample only from the latent posterior. 

We also ran the experiments by integrating over stationary (not input-dependent) signal and noise variances.  However, results with these methods coincide with standard GP regression and the results can be regarded trivial. Thus they are not reported in this paper in order to save space. 

The performance of the different methods was assessed by computing the mean log-predictive density (MLPD) for $N$ test data points
\begin{equation}
	\text{MLPD} = \frac{1}{N} \sum_{i=1}^N \int \log p(y_i^* \mid x_i^*, X, \v y) p(y_i^* \mid x_i^*) \dd y_i^*,
\end{equation}
where $p(y_i^* \mid x_i^*, X, \v y)$ is the posterior predictive density for $y_i^*$ and $p(y_i^* \mid x_i^*)$ is the true distribution of $y_i^*$.
For the three empirical datasets, we computed the approximate MLPD of the $n$ training data points with 10-fold cross-validation:
\begin{equation}
	\text{MLPD} \approx \frac{1}{n} \sum_{i=1}^n \log p(y_i \mid x_i, X_{-i}, \v y_{-i}),
\end{equation}
where $p(y_i \mid x_i, X_{-i}, \v y_{-i})$ is the cross-validated posterior predictive density for $y_i$.  Higher MLPD values correspond to better predictions. 

MLPD values from the experiments are shown in Table~\ref{results}. 
We can conclude from the results that integrating over the input-dependent noise variance increases predictive capability greatly in our experiments compared to standard GP regression. Furthermore, integrating over the input-dependent signal variance tends to enchance the predictions even more. In some cases integration over the signal variance is not needed prediction wise, but our results show that even in these cases, it does not harm the predictive quality. The results show that our EP implementation is comparable to the MCMC methods.  

The predictive distribution with the SP500 data in Figure~\ref{sp500_fig1} illustrates the practical benefits of the input-dependent noise: The period of steady growth between samples 40-80 has clearly lower signal variance compared to the more volatile periods related to financial crisis of 2008 (samples 90-110) and the subsequent shaky growth characterized by debt crises and monetary interventions (samples 110-140). 

With our implementation, MCMC was roughly two orders of magnitude slower than EP. This depends highly on the implementation and number of MCMC draws required for convergence. For example, with the SP500 and Concrete data with ARD lengthscales for $\tilde{f}$, the state-of-the-art MCMC methods based on elliptical slice sampling had convergence issues even after thousands of samples, as the results indicate. 

\section{Discussion}
\label{discussion}

  In this work we have introduced a straightforward but an easily implementable and computationally efficient way to integrate over the uncertainty of the noise and signal variance in Gaussian process regression. Our implementation is easy to apply also for input-dependent noise and signal variance, and it further extends the well-known nonstationary GP models. We have tested our EP implementation on several different data sets and showed that the EP results are on par with state-of-the-art MCMC methods. Furthermore, our results show EP can be used in complex problems where even the state-of-the-art MCMC methods have convergence problems.
  
The scope of this paper was not to compare GPs to other models, but to investigate how integration over signal and noise variance works in the GP framework. Thus, we have ommited comparisons to other models in this work.

The results indicate that there exists phenomena, where it is advantageous to have input dependent signal variance in addition to the input dependent noise variance. While adding the input-dependent noise variance greatly enhances the predictive quality, we are still left with oscillation of the estimated mean. Using the input-dependent signal variance in addition to the noise makes the estimates smoother and further enhances the predictions.

\section{Acknowledgements} 
We acknowledge the computational resources provided by Aalto Science-IT project.

\bibliographystyle{apalike}
\bibliography{paper_epinf_arxiv.bib}

\begin{thebibliography}{}

\bibitem[Adams and Stegle, 2008]{adams2008}
Adams, R.~P. and Stegle, O. (2008).
\newblock Gaussian process product models for nonparametric nonstationarity.
\newblock In {\em Proceedings of the 25th International Conference on Machine
  Learning}, pages 1--8.

\bibitem[Cseke and Heskes, 2011]{cseke2011}
Cseke, B. and Heskes, T. (2011).
\newblock Approximate {M}arginals in {L}atent {G}aussian {M}odels.
\newblock {\em Journal of Machine Learning Research}, 12:417--454.

\bibitem[Diggle and Ribeiro, 2007]{Diggle+Ribeiro:2007}
Diggle, P.~J. and Ribeiro, P.~J. (2007).
\newblock {\em Model-based {G}eostatistics}.
\newblock Springer.

\bibitem[Diggle et~al., 1998]{Diggle+Tawn+Moyeed:1998}
Diggle, P.~J., Tawn, J., and Moyeed, R. (1998).
\newblock Model-based geostatistics.
\newblock {\em Journal of the Royal Statistical Society: Series C (Applied
  Statistics)}, 47(3):299--350.

\bibitem[Gibbs, 1997]{gibbs1997}
Gibbs, M.~N. (1997).
\newblock {\em Bayesian {G}aussian {P}rocesses for {R}egression and
  {C}lassification}.
\newblock PhD thesis.

\bibitem[Goldberg et~al., 1997]{goldberg1997}
Goldberg, P.~W., Williams, C.~K., and Bishop, C.~M. (1997).
\newblock Regression with input-dependent noise: {A} {G}aussian process
  treatment.
\newblock {\em Advances in {N}eural {I}nformation {P}rocessing {S}ystems},
  10:493--499.

\bibitem[Hensman et~al., 2013]{hensman2013}
Hensman, J., Lawrence, N.~D., and Rattray, M. (2013).
\newblock Hierarchical bayesian modelling of gene expression time series across
  irregularly sampled replicates and clusters.
\newblock {\em BMC {B}ioinformatics}, 14(1):1--12.

\bibitem[Jyl{\"a}nki et~al., 2011]{jylanki2011}
Jyl{\"a}nki, P., Vanhatalo, J., and Vehtari, A. (2011).
\newblock Gaussian process regression with a student-t likelihood.
\newblock {\em The Journal of Machine Learning Research}, 12:3227--3257.

\bibitem[Kersting et~al., 2007]{kersting2007}
Kersting, K., Plagemann, C., Pfaff, P., and Burgard, W. (2007).
\newblock Most likely heteroscedastic {G}aussian process regression.
\newblock In {\em Proceedings of the 24th {I}nternational {C}onference on
  {M}achine {L}earning}, pages 393--400.

\bibitem[Le et~al., 2005]{le2005}
Le, Q.~V., Smola, A.~J., and Canu, S. (2005).
\newblock Heteroscedastic {G}aussian process regression.
\newblock In {\em Proceedings of the 22nd {I}nternational {C}onference on
  {M}achine {L}earning}, pages 489--496.

\bibitem[Minka, 2001a]{minka2001a}
Minka, T.~P. (2001a).
\newblock Expectation propagation for approximate {B}ayesian inference.
\newblock {\em Uncertainty in Artificial Intelligence}, 17:362--369.

\bibitem[Minka, 2001b]{minka2001b}
Minka, T.~P. (2001b).
\newblock {\em A {F}amily of {A}lgorithms for {A}pproximate {B}ayesian
  {I}nference}.
\newblock PhD thesis, Massachusetts Institute of Technology.

\bibitem[Naish-Guzman and Holden, 2007]{naish2007}
Naish-Guzman, A. and Holden, S. (2007).
\newblock Robust regression with twinned {G}aussian processes.
\newblock In {\em Advances in Neural Information Processing Systems}, pages
  1065--1072.

\bibitem[Neal, 1998]{neal1998}
Neal, R.~M. (1998).
\newblock Regression and classification using {G}aussian process priors.
\newblock In {\em Bayesian Statistics}, volume~6, pages 475--501. Oxford
  University Press.

\bibitem[Nickisch and Rasmussen, 2008]{nickisch2008}
Nickisch, H. and Rasmussen, C.~E. (2008).
\newblock Approximations for binary {G}aussian process classification.
\newblock {\em Journal of Machine Learning Research}, 9:2035--2078.

\bibitem[Paciorek and Schervish, 2004]{paciorek2004}
Paciorek, C.~J. and Schervish, M.~J. (2004).
\newblock {N}onstationary {C}ovariance {F}unctions for {G}aussian {P}rocess
  {R}egression.
\newblock {\em Advances in neural information processing systems}, 16:273--280.

\bibitem[Rasmussen and Williams, 2006]{gpkirja}
Rasmussen, C.~E. and Williams, C.~K. (2006).
\newblock {\em Gaussian {P}rocesses for {M}achine {L}earning}.
\newblock MIT Press.

\bibitem[Riihim{\"a}ki et~al., 2013]{riihimaki2013}
Riihim{\"a}ki, J., Jyl{\"a}nki, P., and Vehtari, A. (2013).
\newblock Nested {E}xpectation propagation for {G}aussian {P}rocess
  {C}lassification with a {M}ultinomial {P}robit {L}ikelihood.
\newblock {\em Journal of Machine Learning Research}, 14:75--109.

\bibitem[Riihim{\"a}ki and Vehtari, 2014]{riihimaki2014}
Riihim{\"a}ki, J. and Vehtari, A. (2014).
\newblock Laplace approximation for logistic {G}aussian process density
  estimation.
\newblock {\em Bayesian Analysis}, In press.

\bibitem[Silverman, 1985]{silverman1985}
Silverman, B.~W. (1985).
\newblock Some aspects of the spline smoothing approach to non-parametric
  regression curve fitting.
\newblock {\em Journal of the Royal Statistical Society. Series B
  (Methodological)}, pages 1--52.

\bibitem[Vehtari and Lampinen, 2002]{vehtari2002}
Vehtari, A. and Lampinen, J. (2002).
\newblock Bayesian model assessment and comparison using cross-validation
  predictive densities.
\newblock {\em Neural Computation}, 14(10):2439--2468.

\bibitem[Zhang, 2004]{Zhang:2004}
Zhang, H. (2004).
\newblock Inconsistent estimation and asymptotically equal interpolations in
  model-based geostatistics.
\newblock {\em Journal of the American Statistical Association},
  99(465):250--261.

\end{thebibliography}
\end{document}